\pdfoutput=1
\documentclass[10pt,logo,copyright]{nvidiatechreport}
\usepackage[authoryear,sort&compress,round]{natbib}

\usepackage[utf8]{inputenc} % allow utf-8 input
\usepackage[T1]{fontenc}    % use 8-bit T1 fonts

\usepackage{parskip}        % no paragraph indents
\usepackage{url}            % simple URL typesetting
\usepackage{hyperref}       % hyperlinks
\usepackage{booktabs}       % professional-quality tables
\usepackage{amsfonts}       % blackboard math symbols
\usepackage{nicefrac}       % compact symbols for 1/2, etc.
\usepackage{microtype}      % microtypography
\usepackage{xcolor}         % colors
\usepackage[dvipsnames]{xcolor} % more color names
\usepackage{graphicx}
\usepackage{animate}        % for 360 video in teaser
\usepackage{subcaption}
\usepackage{tabularx}
\usepackage{makecell}
\usepackage{adjustbox}
\usepackage{setspace}
\newcolumntype{M}[1]{>{\centering\arraybackslash}m{#1}}
\usepackage{float}
\usepackage{tikz}
\usetikzlibrary{positioning,shapes,arrows}
\usepackage{amsmath,amsfonts,bm, bbm,leftindex}
\usepackage{multirow}
\usepackage{comment}
\usepackage{gensymb}
\usepackage{lipsum}
\usetikzlibrary{arrows.meta, positioning, fit}
\usepackage[para]{threeparttable}
\usepackage{tikz}
\usetikzlibrary{tikzmark}

\def\eqref#1{equation~\ref{#1}}
\def\1{\bm{1}}

\DeclareMathAlphabet{\mathsfit}{\encodingdefault}{\sfdefault}{m}{sl}
\SetMathAlphabet{\mathsfit}{bold}{\encodingdefault}{\sfdefault}{bx}{n}
\makeatletter
\let\save@mathaccent\mathaccent
\newcommand*\if@single[3]{%
  \setbox0\hbox{${\mathaccent"0362{#1}}^H$}%
  \setbox2\hbox{${\mathaccent"0362{\kern0pt#1}}^H$}%
  \ifdim\ht0=\ht2 #3\else #2\fi
  }
\newcommand*\rel@kern[1]{\kern#1\dimexpr\macc@kerna}
\newcommand*\widebar[1]{\@ifnextchar^{{\wide@bar{#1}{0}}}{\wide@bar{#1}{1}}}
\newcommand*\wide@bar[2]{\if@single{#1}{\wide@bar@{#1}{#2}{1}}{\wide@bar@{#1}{#2}{2}}}
\newcommand*\wide@bar@[3]{%
  \begingroup
  \def\mathaccent##1##2{%
    \let\mathaccent\save@mathaccent
    \if#32 \let\macc@nucleus\first@char \fi
    \setbox\z@\hbox{$\macc@style{\macc@nucleus}_{}$}%
    \setbox\tw@\hbox{$\macc@style{\macc@nucleus}{}_{}$}%
    \dimen@\wd\tw@
    \advance\dimen@-\wd\z@
    \divide\dimen@ 3
    \@tempdima\wd\tw@
    \advance\@tempdima-\scriptspace
    \divide\@tempdima 10
    \advance\dimen@-\@tempdima
    \ifdim\dimen@>\z@ \dimen@0pt\fi
    \rel@kern{0.6}\kern-\dimen@
    \if#31
      \overline{\rel@kern{-0.6}\kern\dimen@\macc@nucleus\rel@kern{0.4}\kern\dimen@}%
      \advance\dimen@0.4\dimexpr\macc@kerna
      \let\final@kern#2%
      \ifdim\dimen@<\z@ \let\final@kern1\fi
      \if\final@kern1 \kern-\dimen@\fi
    \else
      \overline{\rel@kern{-0.6}\kern\dimen@#1}%
    \fi
  }%
  \macc@depth\@ne
  \let\math@bgroup\@empty \let\math@egroup\macc@set@skewchar
  \mathsurround\z@ \frozen@everymath{\mathgroup\macc@group\relax}%
  \macc@set@skewchar\relax
  \let\mathaccentV\macc@nested@a
  \if#31
    \macc@nested@a\relax111{#1}%
  \else
    \def\gobble@till@marker##1\endmarker{}%
    \futurelet\first@char\gobble@till@marker#1\endmarker
    \ifcat\noexpand\first@char A\else
      \def\first@char{}%
    \fi
    \macc@nested@a\relax111{\first@char}%
  \fi
  \endgroup
}
\makeatother
\definecolor{darkred}{rgb}{0.7, 0.0, 0.0}

\usepackage{pifont}
\newcommand{\xmark}{\ding{55}} % Cross mark
\usepackage{wrapfig}
\renewcommand{\today}{}
\usepackage[nameinlink]{cleveref}
\crefname{equation}{Eq.}{Eqs.}
\crefname{figure}{Fig.}{Figs.}
\crefname{section}{Sec.}{Sec.}
\crefname{appendix}{App.}{App.}
\crefname{table}{Tab.}{Tabs.}
\crefname{algorithm}{Algo}{Algo}
\crefname{thm}{Thm}{Thm}
\Crefname{thm}{Thm}{Thm}
\crefname{prop}{Prop}{Prop}
\usepackage{longtable}
\usepackage{wrapfig}
\usepackage{caption}
\usepackage{makecell}
\usepackage{xurl}
\usepackage{hyperref}
\usepackage{xcolor}
\usepackage{hyperref} 
\usepackage{tikz}
\usetikzlibrary{trees} % Explicitly load the trees library
\usepackage[edges]{forest}
\usepackage[breakable]{tcolorbox}
\definecolor{nvidiaGreen}{HTML}{9dca63}
\newcommand{\crefnames}[3]{%
  \@for\next:=#1\do{%
    \expandafter\crefname\expandafter{\next}{#2}{#3}%
  }%
}
\usepackage{listings}
\lstdefinestyle{pythonstyle}{language=Python, basicstyle=\ttfamily\small, keywordstyle=\color{blue}, commentstyle=\color{gray}, stringstyle=\color{red}, showstringspaces=false, breaklines=true}

\definecolor{midnightgreen}{rgb}{0.0, 0.29, 0.33}
\definecolor{deepgreen}{HTML}{0aa344}
\definecolor{deeppurple}{HTML}{7030a0}
\definecolor{deepblue}{HTML}{171d91}
\definecolor{brown}{HTML}{843c0c}
\definecolor{shadered}{HTML}{ffe5e5}
\definecolor{shadegreen}{HTML}{e5f7ed}
\definecolor{msftBlack}{RGB}{0,0,0}
\definecolor{lightred}{RGB}{255,163,163}
\definecolor{deepred}{RGB}{146,0,0}
\newtcolorbox{boxL}{
    fontupper = \color{black},
    rounded corners,
    arc = 6pt,
    colframe = black!50, 
    boxrule = 0pt, 
    bottomrule = 4.5pt ,
    breakable,
}
\newcommand{\prorlagent}{\textsc{ProRL Agent }}
\title{ProRL Agent: Rollout-as-a-Service for RL Training of Multi-Turn LLM Agents}
\author{Hao Zhang$^*$, Mingjie Liu$^*$, Shaokun Zhang$^*$, Songyang Han, Jian Hu, Zhenghui Jin, Yuchi Zhang, Shizhe Diao, Ximing Lu, Binfeng Xu, Zhiding Yu, Jan Kautz, Yi Dong
}

\begin{abstract}
Multi-turn LLM agents are increasingly important for solving complex, interactive tasks, and reinforcement learning (RL) is a key ingredient for improving their long-horizon behavior.
However, RL training requires generating large numbers of sandboxed rollout trajectories, and existing infrastructures often couple rollout orchestration with the training loop, making systems hard to migrate and maintain.
Under the \textit{rollout-as-a-service} philosophy, we present \prorlagent, a scalable infrastructure that serves the full agentic rollout lifecycle through an API service.
\prorlagent also provides standardized and extensible sandbox environments that support diverse agentic tasks in rootless HPC settings.
We validate \prorlagent through RL training on software engineering, math, STEM, and coding tasks. \prorlagent is open-sourced at \href{https://github.com/NVIDIA-NeMo/ProRL-Agent-Server}{ProRL Agent} and integrated as part of NVIDIA NeMo Gym.
\end{abstract}

\begin{document}

\maketitle
\abscontent

\section{Introduction}
Recent advances in reinforcement learning from verifiable rewards (RLVR) for large language models (LLMs) are increasingly shifting from single-turn to multi-turn agentic tasks~\citep{guo2025deepseek, hu2025openreasonerzero, cao2025skyrl, luo2025deepswe, gao2025beyond}.
Unlike single-turn tasks, multi-turn agentic tasks typically involves interacting with external environments, such as code repositories~\citep{jimenez2023swe}, web-browser~\citep{zhou2023webarena}, or even full computer operating systems~\citep{xie2024osworld} via iterative tool use. As a result, they often produce trajectories that often span dozens of turns and tens of thousands of tokens.

Training such agents with RL requires repeatedly rolling out policies in these environments and using the resulting trajectories for optimization. As task scale and complexity grow, rollout generation becomes a major bottleneck due to the heterogeneous environments and non-instantaneous feedback inherent in agentic tasks. 
For example, a single rollout in software engineering tasks often involves many sequential environment interactions, each of which may incur highly variable latency depending on the execution result or environment response.
In response, a number of agentic RL training frameworks have recently emerged~\citep{cao2025skyrlagent,jiang2025verltool,tan2025rllm,sheng2025verl,luo2025agentlightning,liu2025gem, xi2026agentgymrl}.

A counterintuitive design in existing frameworks is the tightly coupling agentic rollout with the RL training stack, with agent lifecycle handled within the trainer. This couples two modules with fundamentally different responsibilities leads to two major limitations.
\begin{enumerate}
    \item \textbf{Conflicting system requirements:} Rollout and policy training have fundamentally different resource and operational characteristics. Rollout is I/O-intensive, involving sandbox creation, long-lived tool sessions, and asynchronous coordination across hundreds of concurrent instances. Training, by contrast, is GPU-intensive, centered on forward and backward passes, and gradient synchronization. Coupling these workloads causes interference and reduces overall resource efficiency.
    \item \textbf{Difficult to migrate and maintain:} When rollout logic is embedded in RL trainer, migrating to a different training backend often requires re-implementing the entire agent execution pipeline. Likewise, improving the rollout infrastructure, such as supporting new runtime environments or tasks, often requires changes that propagate into the training codebase. In practice, this tight coupling slows progress on both fronts, as it makes independent experimentation and optimization on either side more difficult.
\end{enumerate}
These issues are likely to be further exacerbated by the growing need for rapid infrastructure iteration and more effective use of compute resources. If rollout and training are not decoupled from the begining, the accumulated system complexity can become a serious obstacle to scalability and long-term maintainability.

Drawing inspiration from the inference-as-a-service philosophy adopted by common LLM inference engines~\citep{kwon2025vllm, zheng2024sglang}, we adopt \texttt{rollout-as-a-service} as the core design principle for agentic RL training frameworks, decoupling the trainer from agentic rollout by treating the agentic rollout lifecycle as an independent service. 
We present \prorlagent, an open-source scalable infrastructure for multi-turn agentic rollout in RL training. 
Instead of implementing rollout as an in-process component of the RL trainer, \prorlagent serves the full rollout pipeline, from environment initialization to outcome evaluation, through an HTTP server. 
This design allows RL trainers to submit task instances and retrieve completed trajectories without managing any part of the rollout lifecycle. 
On one hand, this decoupled design allows rollout and training to run on different machines, separating I/O-intensive execution from GPU-intensive optimization; on the other hand, it improves extensibility and maintainability by decoupling rollout infrastructure from training backends.

In addition, \prorlagent provides several other features that support effective RL training for multi-turn agents. 
First, it adopts token-in/token-out communication throughout the training pipeline, allowing trainers to directly consume token-level trajectories while avoiding re-tokenization drift~\citep{team2025notokenization}. This makes training more stable and faithful to the original model outputs.
Second, \prorlagent provides extensible sandbox environments for agent execution, with flexible support for diverse tools and task. This makes it simple to host heterogeneous agentic tasks within a unified rollout service.
Third, \prorlagent is designed for rootless deployment in shared cluster environments. This makes it practical to run large-scale agentic rollouts under the permission and isolation constraints common in HPC settings.

\begin{figure}[t]
  \centering
  \includegraphics[width=1.0\linewidth]{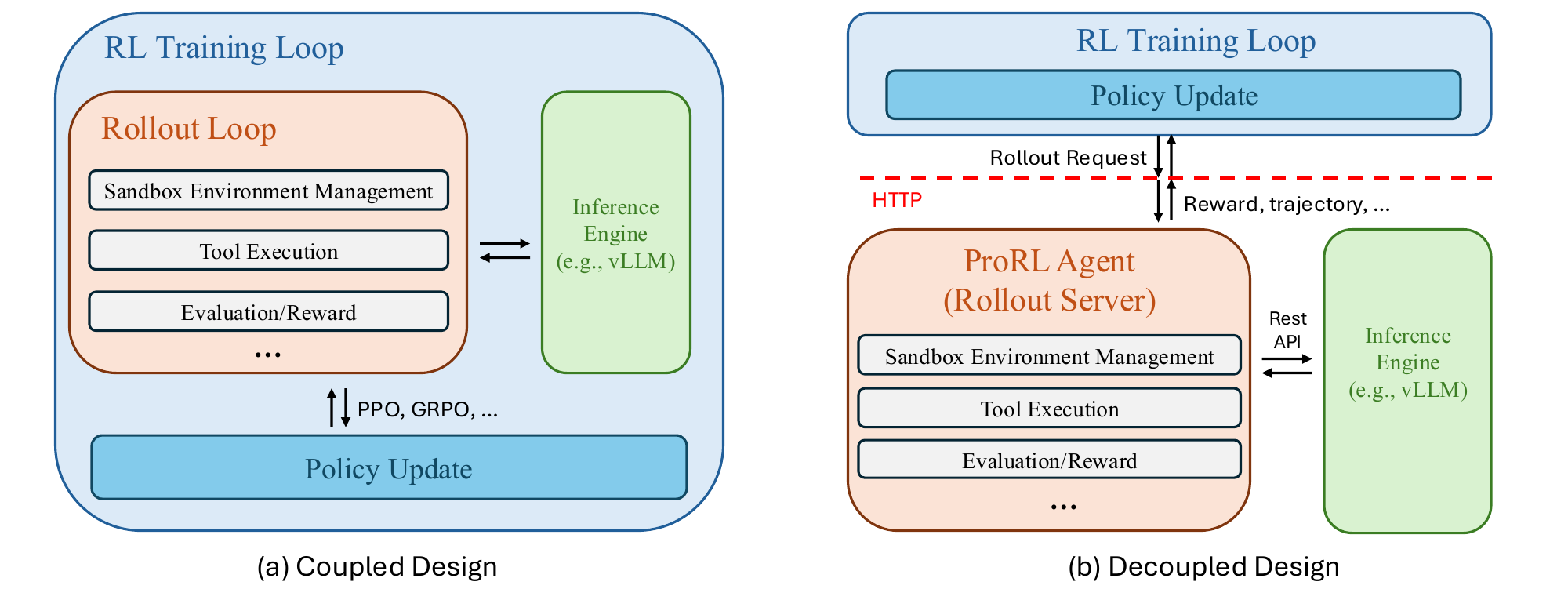}
\caption{\textbf{Coupled vs.\ decoupled designs.}
\textbf{Left:} Existing frameworks often embed the full agentic rollout lifecycle inside the RL training stack. 
\textbf{Right:} \prorlagent treats rollout as an independent HTTP service. The trainer submits rollout requests and receives completed trajectories and rewards, while the rollout server handles environment execution, tool use, evaluation, and inference coordination. This decoupled design improves resource isolation, portability, and extensibility.}
  \label{fig:decoupling}
  \label{fig:teaser}
\end{figure}

We validate \prorlagent by integrating it with ProRL training framework~\citep{prorl2025} for end-to-end RL training on software engineering tasks. Across 4B, 8B, and 14B model scales, it yields strong gains on SWE-Bench Verified. It also performs well in other agentic domains, including MATH, STEM, and coding. ProRL Agent is also integrated as part of NVIDIA NeMo Gym~\citep{nemo-gym}.

In summary, the main contributions of this work are:
\begin{itemize}
    \item We identify the key limitation in existing agentic RL training frameworks: multi-turn agentic rollout is typically tightly coupled with the RL training stack, even though rollout and training have fundamentally different resource and execution characteristics. To address this, we introduce ProRL Agent, an open-source and scalable rollout infrastructure for agent RL training built on the \texttt{rollout-as-a-service principle}, which decouples the full rollout lifecycle from the trainer through a unified HTTP interface.
    \item We design ProRL Agent with several practical properties for multi-turn RL training, including token-in/token-out trajectory communication to avoid re-tokenization drift, extensible sandboxed environments for heterogeneous tools and tasks, and rootless deployment support for shared HPC clusters.
    \item We validate ProRL Agent through end-to-end RL training on software engineering tasks with the ProRL training framework. Across 4B, 8B, and 14B model scales, it achieves strong gains on SWE-Bench Verified, while also showing strong performance in other agentic domains such as math, STEM, and coding.
\end{itemize}

\section{Related Work}
\begin{table*}[h]
\centering
\small
\setlength{\tabcolsep}{3pt}
\begin{tabular}{lccc}
\toprule
\textbf{Frameworks} & \textbf{Training-Rollout Decoupled?} & \textbf{Rootless Sandbox?} & \textbf{Scaffold-Independent?} \\
\midrule
SkyRL-Agent~\citep{cao2025skyrlagent} & \xmark & \xmark & \checkmark \\
VeRL-Tool~\citep{jiang2025verltool} & \xmark & \xmark & \checkmark \\
Agent Lightning~\citep{luo2025agentlightning} & \xmark & \xmark & \xmark \\
rLLM~\citep{tan2025rllm} & \xmark & \xmark & \checkmark \\
GEM~\citep{liu2025gem} & \xmark & \xmark & \checkmark \\
\midrule
\rowcolor{gray!30} \textbf{\prorlagent (Ours)} & \checkmark & \checkmark & \checkmark \\
\bottomrule
\end{tabular}
\caption{\textbf{Comparison of \prorlagent with existing frameworks for multi-turn agent RL.} \prorlagent decouples rollout from training, supports rootless sandboxing for shared HPC environments, and is independent of any specific training framework.}
\label{tab:framework_comparison}
\end{table*}

\textbf{Multi-turn RL for LLM Agents.}
Reinforcement learning has been highly effective for improving single-turn reasoning such as mathematics, logic, and coding~\citep{shao2024deepseekmath, guo2025deepseek, hu2025openreasonerzero, zhang2026nemotronresearchtooln}. 
Building on this progress, recent work has extended RL to multi-turn agentic settings, where agents interact with external environments over long horizons~\citep{cao2025skyrl, luo2025deepswe, gao2025beyond, li2025torl, jin2025searchr1,wang2025vagen,ragenv2026collapse}. 
In these settings, a multi-turn agent is naturally formulated as a POMDP~\citep{kaelbling1998pomdp}, where agent produces actions through tool calls~\citep{yao2022react, wang2024codeact,patil2025the,zhang2024offline} and receives environment observations at each step.
As tasks become more complex, multi-turn rollouts often span dozens of steps in diverse environments, such as code repositories~\citep{jimenez2024swebench, jain2025r2egym}, web browsers~\citep{zhou2023webarena}, and even computer operating systems~\citep{xie2024osworld}.

As a result, the infrastructure required to generate, manage, and evaluate these rollouts at scale has become a major bottleneck for RL training.
This bottleneck slows both training and the deployment of RL agents. \prorlagent is designed to address this challenge by decoupling the full lifecycle of multi-turn agent rollout from the training stack, allowing researchers and practitioners to focus on training algorithms and agent design.

\textbf{Agent RL Infrastructures.}
A growing body of work has begun to address the challenges of scalable RL training for agents, including support for diverse tool integration~\citep{jiang2025verltool, li2025torl}, flexible environment abstractions~\citep{liu2025gem, tan2025rllm}, and efficient rollout scheduling~\citep{cao2025skyrlagent}.
Yet across these frameworks, rollout orchestration, including environment lifecycle management, tool execution, trajectory collection, and evaluation, remains implemented as an in-process library within the training loop. 
Under this design, adopting a new training backend often requires re-implementing or porting the entire rollout stack. This tight coupling makes rollout infrastructure a major source of friction in multi-turn agent RL, often demanding more engineering effort than the training algorithm itself.

\textbf{Agentic Sandbox Environments.}
Multi-turn agent training requires sandboxed environments that provide isolation, reproducibility, and security at scale.
Existing platforms~\citep{wang2024openhands, jimenez2024swebench, jain2025r2egym, yang2024sweagent} have established primary protocols, but they deeply rely on Docker for agent execution.
Docker assumes daemon access and root-equivalent privileges, which are often unavailable on shared Slurm-managed HPC clusters. 
As a result, practitioners often face a trade-off between maintaining separate infrastructure for evaluation and deployment, or incurring the operational complexity of privileged container runtimes on restricted systems.
\prorlagent addresses this limitation by building its sandbox infrastructure on Singularity, enabling rootless execution and native Slurm integration for large-scale agent training on HPC systems.

\section{System Design: Training–Rollout Decoupling}

\begin{figure}[htb]
  \centering
  \includegraphics[width=1.0\linewidth]{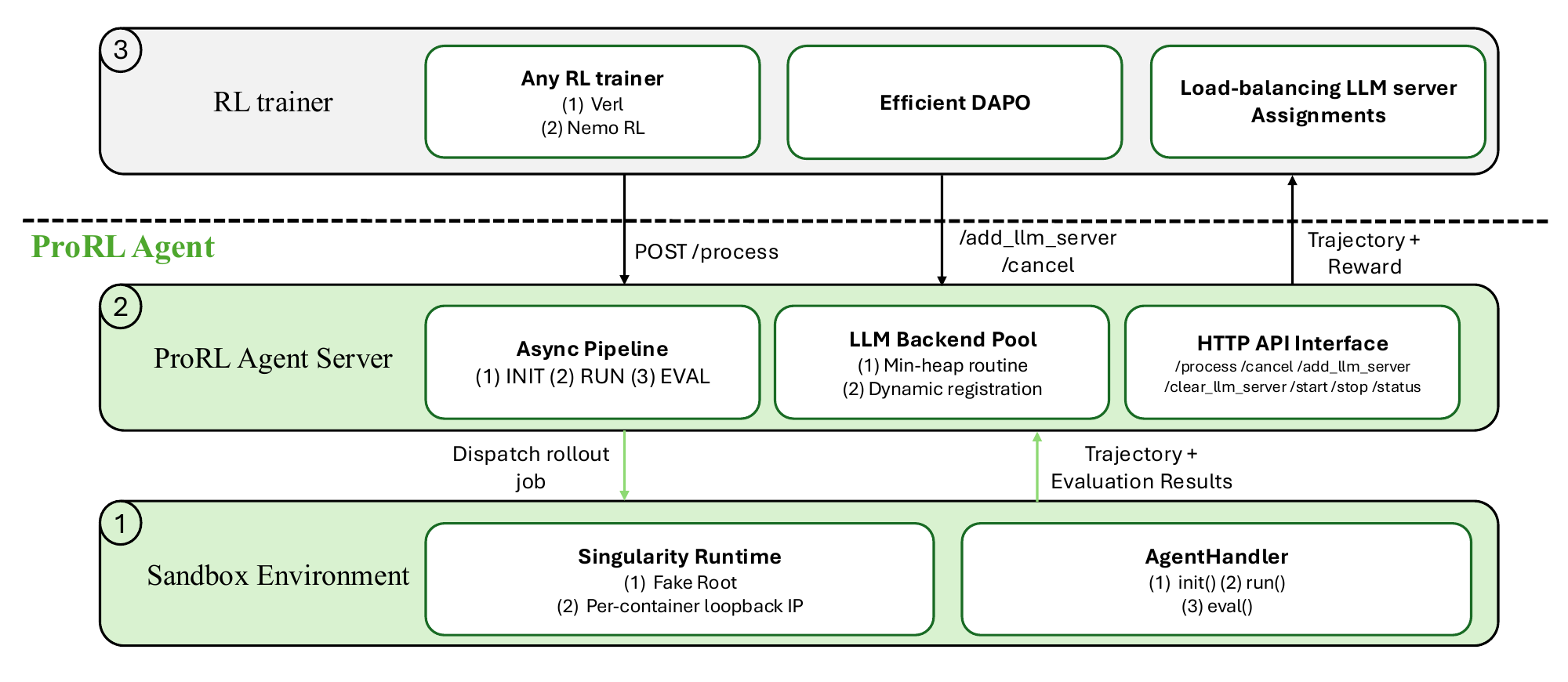}
  \caption{Overview of the \prorlagent architecture. The system consists of three components. 
  \textbf{(1) Sandbox Environment}: each rollout is executed inside a \texttt{SingularityRuntime} container and orchestrated via \texttt{AgentHandler}, which exposes three lifecycle methods including \texttt{init()}, \texttt{run()}, and \texttt{eval()} for environment setup, multi-turn agent execution, and reward scoring, respectively. 
  \textbf{(2) ProRL Agent Server}: an HTTP service that manages rollouts through a three-stage asynchronous pipeline (INIT~$\to$~RUN~$\to$~EVAL) with independent worker pools, and maintains a min-heap LLM backend pool supporting dynamic registration and checkpoint swapping. 
  \textbf{(3) RL Trainer}: any training framework (e.g., veRL, NeMo RL) interacts with the server solely via HTTP, submitting jobs via \texttt{POST \/process} and managing backends via \texttt{\/add\_llm\_server} and \texttt{/cancel}; completed trajectories and rewards are returned to the trainer to update the policy.}
  \label{fig:system_design}
\end{figure}

\subsection{Overview}
\label{sec:overview}

Training RL agents on agentic tasks normally involves multi-turn interaction with live execution environments, where each data sample spans sandbox environment setup, tool execution, and outcome scoring, a process far more complex than single-step generation.
Prior systems typically embed rollout logic directly inside the training loop~\citep{cao2025skyrlagent}, tightly coupling the agent task loop, execution environment, and RL algorithm.
This coupling imposes significant engineering overhead when switching task, and RL trainers.

\prorlagent addresses this through a rollout-as-a-service design with \textbf{rollout-level decoupling}, in which rollout orchestration is fully separated from the training process.
In particular, ProRL Agent Server runs as a standalone HTTP service that accepts a task instance, executes the full agent rollout internally, and returns a completed trajectory with a reward signal. The training framework interacts with the server only through this interface, remaining agnostic to RL infrastructure. This decoupling has three practical consequences.
\begin{itemize}
    \item The RL trainer and agentic rollout logic can be developed, deployed independently: rollout nodes and training nodes can be optimized seperately for larger throughput.
    \item Adding a new task requires only implementing a handler plugin on the rollout server side, with no changes to the training code.
    \item Agentic scaffolds can be modified or replaced without affecting the training infrastructure, as the rollout service and the agent implementation are fully decoupled. 
\end{itemize}
Figure~\ref{fig:system_design} illustrates the overall architecture, which consists of three main components: extensible sandbox environments, the ProRL Agent server for rollout scheduling, and the RL training backend. We introduce each component in turn and describe how they interact within the system.

\subsection{Extensible Sandbox Environments} 
\label{sec:sandbox}

Performing RL training over diverse multi-turn agentic tasks normally requires a sandbox layer that can accommodate heterogeneous task environments and run portably on HPC clusters without privileged access.
We build such the sandbox system around two components: a pluggable task abstraction that decouples task-specific logic from the server core, and an HPC-compatible container runtime that enables isolated, rootless agentic tasks execution at scale.

\subsubsection{Pluggable Task Abstraction}
\label{sec:handlers}

Different agentic tasks e.g., software engineering, mathematical reasoning, computer use, each require their own environment setup, agent behavior, and reward computation.
Hardcoding these differences in the server would make it brittle and always rely on great human efforts.
Instead, we encapsulate all task-specific logic in an abstract interface called \emph{AgentHandler}, which defines three core lifecycle methods corresponding to the three pipeline stages:

\begin{itemize}
  \item \texttt{init}: initialize the sandbox environment for the task, configures the agent with corresponding toolset.
  \item \texttt{run}: drives the multi-turn agent loop within the prepared sandbox environment, collecting the action-observation trajectory and any task artifacts.
  \item \texttt{eval}: scores the agent's output against the ground truth and returns a scalar reward signal for subsequent RL training.
\end{itemize}

Each handler additionally exposes per-stage error callbacks (\texttt{init\_exception}, \texttt{run\_exception}, \texttt{eval\_exception}) and a \texttt{final\_result} method for response serialization, ensuring the server always emits a well-formed output even when a rollout fails partway through.
Listing~\ref{lst:handler} illustrates the interface and a minimal registration example.

\begin{lstlisting}[
  style=pythonstyle,
  caption={The \texttt{AgentHandler} interface and task registration. Each task domain subclasses \texttt{AgentHandler} and registers under a unique name. The server dispatches incoming jobs by matching with different registry.},
  label={lst:handler}
]
class AgentHandler(ABC):
    @abstractmethod
    async def init(self, job_details) -> (Runtime, Metadata, Config):
        """Provision environment; return (runtime, metadata, config)."""
    @abstractmethod
    async def run(self, job_details) -> dict:
        """Execute agent loop; return trajectory and artifacts."""
    @abstractmethod
    async def eval(self, job_details) -> dict:
        """Score output; return reward signal."""
    # Error callbacks (one per stage) and result serializer
    def init_exception(self, job_details, exc) -> dict: ...
    def run_exception(self, job_details, exc)  -> dict: ...
    def eval_exception(self, job_details, exc) -> dict: ...
    def final_result(self, job_details)        -> dict: ...
\end{lstlisting}

When the server receives a job, it reads the task instance, looks up the corresponding handler in the registry, and dispatches to its lifecycle methods in order.

\subsubsection{HPC-Compatible Container Runtime}
\label{sec:runtime}

Most agentic sandbox environments assume a cloud or workstation environment where Docker is readily available.
HPC clusters, however, typically forbid Docker daemons for security reasons, requiring all user processes to run without root privileges under a batch scheduler such as Slurm.
To bridge this gap, we implement \emph{SingularityRuntime}, a container system that requires no persistent daemon and runs entirely as an unprivileged user process to serve sandbox environments.  

\textbf{Container isolation and port management.} 
Each container is launched as a child process in its own session; shutdown proceeds gracefully via \texttt{SIGTERM} before escalating to \texttt{SIGKILL} if necessary.
To support many concurrently running containers on the same node without port conflicts, each container instance is assigned a unique loopback IP address within the \texttt{127.x.x.x} range via a thread-safe allocator.
Two flags address common HPC constraints: \texttt{--fakeroot} grants the container simulated root access for package installation without requiring actual host privileges, and \texttt{--network none} optionally disables external network access to isolate rollouts from interference.

\textbf{Image build pipeline.} 
Container images are packaged as Singularity Image Files (.sif), which encapsulate the full execution environment in a single portable file. This format is particularly well-suited to Slurm shared filesystems, where no persistent container daemon is available.
A companion \texttt{SingularityRuntimeBuilder} constructs images from Jinja2 templates and supports three caching modes: 
\textsc{Scratch} always performs a full rebuild; \textsc{Versioned} reuses a cached image when the base image and framework version are unchanged; and \textsc{Lock} reuses it whenever the dependency lockfile is identical.
The template-driven design enables flexible specialization of runtimes for heterogeneous agentic environments.
For example, QEMU-based virtual machines used in GUI-centric tasks can provide custom definition files to the builder without requiring any modifications to the core build logic.

\subsubsection{Efficient tool backends}
The agent mostly interacts with the environment through tools: it reads and writes files, executes shell commands, runs Python code, and browses the web.
Each tool call is a synchronous blocking operation from the agent's perspective, the agent must wait for the observation before it can decide its next action.
Because a typical rollout spans dozens of such calls, per-tool latency compounds directly into total rollout time, and at high concurrency this overhead can dominate LLM inference as the primary bottleneck.
We therefore optimize three critical tool backends.

\textbf{Efficient Bash.}
Shell execution is the most frequent action across all code-centric agentic tasks. 
Conventional implementations route bash commands through a tmux session, incurring the overhead of terminal multiplexing.
We replace this with a \texttt{ptyprocess}-based direct pseudo-terminal, which grants the agent a raw shell without the tmux intermediary, yielding a significant reduction in shell command round-trip latency.

\textbf{IPython.}
When an agent writes and executes Python code across multiple steps, it is often building on its own prior work: importing a library once, then using it repeatedly; defining a helper function, then calling it later.
A persistent IPython kernel makes this natural so that variables and imports defined in one step remain available in subsequent steps, so the agent does not need to repeat setup code on every call.
The conventional way to host such a kernel is through the Jupyter kernel gateway, but this adds a network round-trip even when the kernel runs on the same machine as the agent.
We instead connect to the kernel directly via its in-process API, removing this overhead entirely. 

\textbf{UDS communication.}
When the agent decides to take an action, such as running a shell command, editing a file, or executing Python, that action is not run directly by the agent process.
Instead, it is sent to a small execution server running inside the container, which carries out the action and sends the observation back.
The common transport for this channel is TCP loopback, which works correctly but forces co-located processes that share the same IP to be distinguished only by port numbers, complicating non-conflicting port assignment and it typically offers lower throughput than Unix domain sockets.
We replace it with Unix domain sockets (UDS), a simpler IPC mechanism that passes messages through the OS kernel directly without any networking overhead. Since this channel is exercised on every agent action, shaving latency here accumulates meaningfully across a full rollout. 

Together, these three optimizations ensure that tool execution does not become the throughput bottleneck as rollout concurrency scales to hundreds of parallel agents.

\subsection{ProRL Agent Server}
\label{sec:server}

With the sandbox layer handling individual rollout execution, the server's
core responsibility during RL training is to orchestrate hundreds of such
rollouts concurrently while providing the training framework with live control
over the rollout infrastructure.

There are two basic requirements for the server:
\begin{itemize}
    \item First, the three rollout phases have fundamentally different resource demands: container initialization is I/O-bound, agent execution is LLM-inference-bound, and outcome evaluation ranges from a few milliseconds for direct scoring to several minutes for full test-suite execution. Executing these phases within each job should not limit throughput to the slowest stage.
    \item Second, the training framework needs dynamic control over LLM inference backends: it must be able to register new servers as the compute cluster scales, swap backends when model checkpoints are updated, and cancel stale in-flight jobs whose gradient batch has already advanced, all without tight coupling to the server internals. 
\end{itemize}

ProRL Agent Server addresses both facets through two mechanisms: 
\textbf{(1) An asynchronous three-stage pipeline} that assigns each rollout phase to an independent worker pool so all three phases can overlap across the job population; and 
\textbf{(2) A lightweight management API} that exposes job submission, per-job cancellation, LLM backend registration, and server lifecycle control to any RL training framework over HTTP.
Listing~\ref{lst:server} sketches the resulting architecture.

\begin{lstlisting}[
  style=pythonstyle,
  label={lst:server},
  caption={Simplified logic of the ProRL Agent Server.  Three independent
           worker pools drain their respective queues concurrently.},
]
# -- Three independent worker pools -------------------------------
STAGES = [INIT, RUN, EVAL]
queues = {s: Queue() for s in STAGES}   # thread-safe FIFO per stage
pools  = {s: ThreadPool(N[s]) for s in STAGES}
llm_backends = MinHeap()                # min-heap keyed by in-flight count

def worker_loop(stage):
    while running:
        job = queues[stage].get()
        if job.id in discarded: continue
        with job.timer.phase(stage):    # only this phase counts toward timeout
            try:
                result = handler[stage](job)
            except Exception as e:
                result = handler[stage+'_exception'](job, e)
        job.store(stage, result)
        if stage == RUN:
            cleanup(job.runtime)        # free container before eval starts
        if stage != EVAL:
            queues[next_stage[stage]].put(job)
        else:
            job.done.set()              # unblock the waiting HTTP handler
\end{lstlisting} 

\subsubsection{Three-Stage Rollout Pipeline}
\label{sec:pipeline}

Think of the rollout process as an assembly line. A naive implementation would assign one worker to each job and have that worker do everything: start the container, run the agent, and score the result, before picking up the next job.
The problem is that each phase takes a very different amount of time and uses a very different resource.
Container startup is slow because it is waiting on disk I/O and the network.  Agent execution is fast per call but fires dozens of LLM requests, so it is bottlenecked by GPU throughput.
Evaluation can be nearly instant for a math answer check, or take several minutes for a full test suite.
A single worker sitting through all three phases in sequence would spend most of its time idle, waiting for whichever phase happens to be slow.

In ProRL Agent server, the solution is to decouple the phases, exactly as a factory decouples assembly stations.
The three lifecycle methods of \texttt{AgentHandler}(Section~\ref{sec:handlers}) map onto three independent worker pools, each with its own queue.
Initialization workers continuously pull new jobs, spin up containers, and hand them off to the rollout queue.
Rollout workers drive agent loops and hand completed trajectories to the evaluation queue.
Evaluation workers score results and return them to the caller. 

At any moment, all three pools are busy on different jobs simultaneously: while one job is being evaluated, a second is mid-rollout, and a third is having its container started.
Because the pools are independent, they can also be sized separately to match their respective workloads, with more init workers to absorb the slow I/O startup, or more eval workers when test suites are particularly long.

\subsubsection{LLM Backend Management}
\label{sec:llm-mgmt}

\begin{lstlisting}[
  style=pythonstyle,
  label={lst:server_api},
  caption={Simplified logic of the ProRL Agent Server. The
           management API gives the training framework full control over jobs
           and LLM backends at runtime.},
]
# -- Management API (HTTP endpoints) -------------------------------
POST /add_llm_server   {"address": "http://host:port/v1"}  # register backend
POST /clear_llm_server                                      # flush all backends
POST /process  {"instance": {...}, "sampling_params": {...}}  # submit job
POST /cancel   {"job_id": "..."}                            # abort running job
POST /start  |  POST /stop                                  # server lifecycle
GET  /status                                                # queue depths
\end{lstlisting}

Every step of the agent loop requires an LLM completion: the model receives the current conversation history and produces the next action.
When hundreds of rollouts run in parallel, these calls arrive at the inference layer simultaneously and at high frequency.
A single LLM server (e.g., vLLM server) quickly becomes a bottleneck, so RL training typically co-deploys a pool of LLM servers and distributes inference traffic across them.
The ProRL Agent Server manages this pool directly, handling both registration and routing so that the training framework does not need to coordinate LLM access itself.

\textbf{Dynamic registration and checkpoint swapping.}
LLM backends are registered and deregistered through the management API at any time during a training run. We show the simplified logic in Listing~\ref{sec:llm-mgmt}
When a new LLM server comes online, the trainer calls \texttt{POST /add\_llm\_server} with the server's endpoint; the server is immediately available for routing.
When the RL trainer updates the policy checkpoint (e.g., after a gradient synchronization step), the old LLM weights are no longer valid.
Rather than restarting the rollout server, the trainer calls \texttt{POST /clear\_llm\_server} to flush all registered backends, then re-registers the reloaded LLM server endpoints.
From that point on, all subsequent rollouts automatically use the updated model, with no interruption to jobs already in the pipeline.

\textbf{Load balancing via min-heap.}
Each LLM backend is stored alongside an assignment counter in a min-heap.
Every time the rollout stage needs to issue an LLM call, ProRL Agent server automatically selects the backend with the lowest counter and assigns that entire task to the selected LLM. The counter is incremented once per task (rather than per call), ensuring that all subsequent calls within the same task are consistently routed to the same backend to maximize prefix cache reuse. After assignment, the backend’s updated counter is used to maintain its position in the heap:
\begin{align*}
  s^{*} = \arg\min_{s}\; w_{s}, \qquad w_{s^{*}} \leftarrow w_{s^{*}} + 1,
\end{align*}
where $w_s$ counts the total number of inference calls assigned to server $s$ since it was registered.
Because selection is proportional to assignment count, servers that receive heavier traffic fall back in priority, achieving a round-robin-like balance across the pool without requiring any global synchronization.
The entire operation is protected by a single lock, making it safe under the high concurrency of the rollout worker pool.

\subsubsection{Token-in/Token-out}
If trajectories are transmitted through the training pipeline as plain text, re-tokenization on the can be lossy: the resulting token sequence may differ from the one originally generated during rollout~\citep{team2025notokenization}, leading to unintended off-policy discrepancies.

\prorlagent eliminates this \emph{re-tokenization drift} by using token IDs as the canonical representation throughout the entire training process.
The rollout worker sends \texttt{prompt\_ids} directly to the LLM backend and receives \texttt{response\_ids} with per-token log-probabilities; each message additionally carries \texttt{input\_ids}, \texttt{output\_ids}, and \texttt{logprobs} fields that are populated at generation time and propagated unchanged.
During multi-turn rollouts, prior assistant turns retain their original token IDs and are concatenated directly into the input buffer; only new messages (e.g., environment observations) are tokenized and appended.
This ensures that every token ID returned to the trainer is identical to the one produced during rollout.

\subsubsection{Job Lifecycle and Cancellation}
\label{sec:lifecycle}

We then describe the lifecycle of each job instance and the cancellation mechanism, which together provide greater flexibility for RL trainers.

\textbf{Phase-aware timeouts.}
Each job is associated with a \texttt{PausableTimer} that accumulates elapsed time only during active pipeline stages (\textit{init}, \textit{run}, and \textit{eval}), while excluding time spent waiting in inter-stage queues. This design ensures that the timeout budget reflects actual execution time rather than transient server-side delays.

\textbf{Cancellation.}
The training framework can abort any in-flight job at any time via \texttt{POST /cancel}.
Once received, ProRL Agent server will: 
(i)mark the job as discarded so that any worker that has not yet dequeued it will skip it; 
(ii)cancel the currently executing async task; 
(iii)close the associated container runtime to release resources immediately; and 
(iv)signal the job's completion event so the waiting HTTP handler returns without blocking.
This enables the RL trainer to discard incomplete rollouts once a sufficient number of valid samples has been collected.

\textbf{Fault isolation.}
Each pipeline stage registers a dedicated exception callback. Once failure, the callback populates \texttt{JobDetails} with a structured fallback result and sets the completion event, preventing any single failed rollout from stalling the shared worker pool.

\textbf{Graceful shutdown.}
Once received \texttt{POST /stop}, the server cancels all in-flight jobs, terminates Singularity processes via process-group scanning, drains the worker pools, and exits cleanly, leaving no orphaned containers on the node.

\begin{figure}[t]
  \centering
  \includegraphics[width=1.0\linewidth]{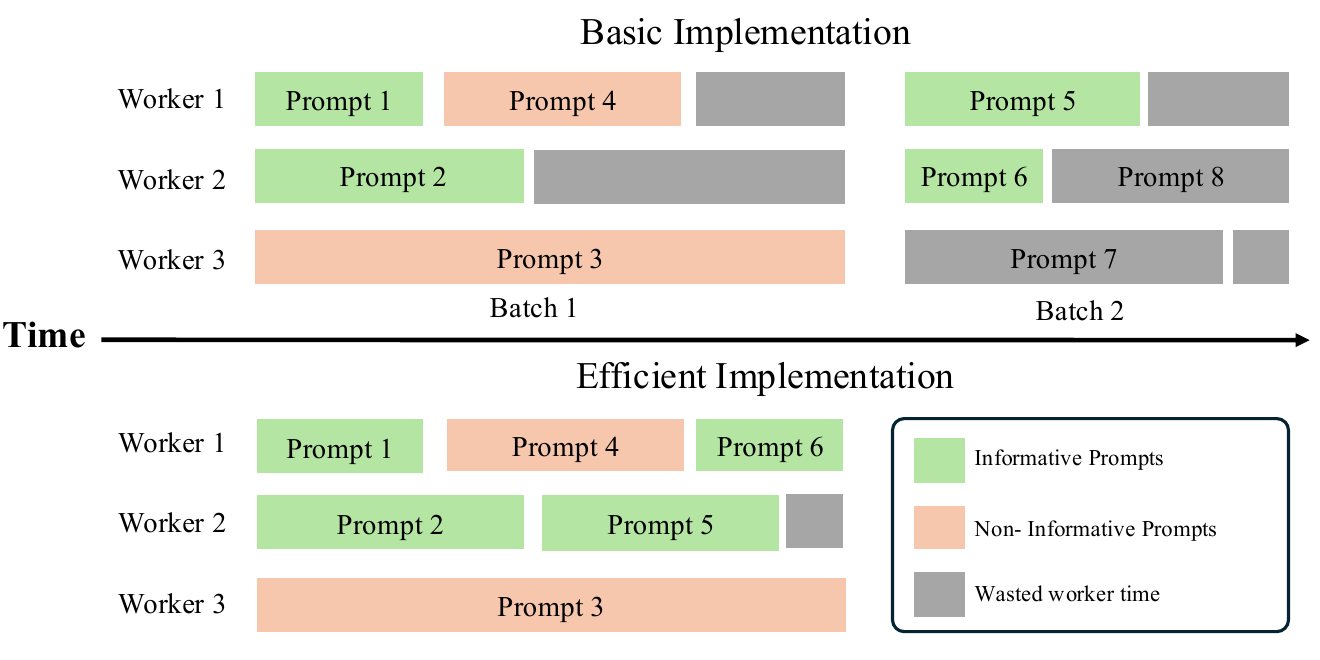}
\caption{Comparison of DAPO implementations ($n=4$). Our efficient implementation optimizes worker synchronization, significantly reducing the idle time (waiting period) between rollout generations compared to the baseline batch-by-batch approach.}
  \label{fig:dapo}
\end{figure}

\subsection{Connecting to RL Trainers}
Rollout-level decoupling allows the agent server to interface with a wide range of RL trainers. In our implementation, we support both VeRL~\cite{sheng2025verl} and NeMo RL~\cite{nemo-rl}. In addition, we provide several key features that further improve RL training.

\textbf{Efficient Asynchronous Task Scheduling.} On the RL client side, we implement a two-phase hierarchical load balancing strategy that jointly optimizes communication locality and global load balance. In the first phase, LLM servers are assigned preferentially to \prorlagent servers on the same physical node, identified through IP address matching, to reduce network latency. In the second phase, any remaining servers are distributed in a round-robin manner to maintain balanced allocation across all available LLM servers.

\textbf{Efficient DAPO.} We adopt Dynamic Sampling Policy Optimization (DAPO)~\citep{yu2025dapoopensourcellmreinforcement} as our core reinforcement learning algorithm. DAPO enhances training stability and data efficiency by filtering out Zero-Variance Prompts—those whose rollouts yield uniform rewards (e.g., all correct or all incorrect) and thus provide no gradient signal. However, applying DAPO to Agent RL is challenging because agent rollouts are typically long-running, asynchronous, and computationally expensive.A naive batch-by-batch implementation—where the trainer requests $n$ prompts, filters out the non-informative ones, and repeatedly triggers new batches until $n$ Informative Prompts are collected—is highly inefficient. This synchronous approach leads to worker idle time and generates redundant rollouts that exceed the target count. Furthermore, discarding incomplete rollouts at the end of a batch results in significant data waste.To address these bottlenecks, we implement an asynchronous replenishment mechanism:
\begin{enumerate}
    \item \textbf{Continuous Throughput:} We replenish the job queue as soon as it empties to maintain maximum rollout throughput.
    \item \textbf{Early Termination:} We terminate remaining active jobs once the target number of Informative Prompts is reached.
    \item \textbf{Cross-Iteration Persistence:} Unfinished jobs are carried over to the subsequent iteration to preserve partial progress.
    
\end{enumerate}
As illustrated in \Cref{fig:dapo}, our optimized implementation significantly reduces worker idle time and improves overall hardware utilization compared to the baseline.

\section{Experiments}
We next present the experimental results of \prorlagent across different tasks. We also perform in-depth investigations to provide a better understanding of our infrastructure.

\subsection{Experimental Setup}

Unless otherwise specified, we adopt DAPO~\citep{yu2025dapoopensourcellmreinforcement} as the default RL algorithm which filters out instances that are either too easy (resolved ratio 100\%) or too hard (resolved ratio 0\%).
We use a batch size of 32, a mini-batch size of 8, and generate 8 rollouts per instance. 
Rollouts with errors are excluded from gradient computation. The KL coefficient is set to $1\times10^{-4}$ and the learning rate to $1\times10^{-6}$. All RL training is performed on 32 NVIDIA H100 GPUs. 

\begin{table}[htb]
\centering
\small
\caption{Comparison of performance on SWE-Bench Verified across models of different scales. We report the reproduced performance and, where available, the reported results from prior work. Across all model sizes}
\label{tab:swe_results}
\setlength{\tabcolsep}{6pt}
\renewcommand{\arraystretch}{1.1}
\begin{tabular*}{\linewidth}{@{\extracolsep{\fill}}llcc@{}}
\toprule
\textbf{Size} & \textbf{Model} & \textbf{Reproduced} & \textbf{Reported} \\
\midrule
\multirow{2}{*}{4B} 
& Qwen3-4B-Instruct-2507        & 14.8 & --   \\
& \cellcolor{gray!30} \textbf{ProRL Agent-4B (RL)}  & \cellcolor{gray!30}\textbf{21.2} & \cellcolor{gray!30}--   \\
\midrule
\multirow{3}{*}{8B} 
& Qwen3-8B                      & 9.6  & --   \\
& SkyRL-Agent-8B-v0             & --   & 9.4  \\
& \cellcolor{gray!30}\textbf{ProRL Agent-8B (RL)}  & \cellcolor{gray!30}\textbf{18.0} & \cellcolor{gray!30} --   \\
\midrule
\multirow{3}{*}{14B} 
& Qwen3-14B                     & 15.4 & --   \\
& SkyRL-Agent-14B-v0            & --   & 21.6 \\
& \cellcolor{gray!30} \textbf{ProRL Agent-14B (RL)} & \cellcolor{gray!30} \textbf{23.6} & \cellcolor{gray!30} --   \\
\bottomrule
\end{tabular*}
\end{table}

\subsection{Main Results on Software Engineering}

We primarily evaluate \prorlagent on software engineering tasks. Specifically, we train Qwen3-4B-Instruct-2507, Qwen3-8B, and Qwen3-14B on the 293-instance subset of SWE-Gym used in SkyRL-v0~\citep{cao2025skyrl}. For the thinking models, Qwen3-8B and Qwen3-14B, we enable thinking mode during training. The results are reported in Table~\ref{tab:swe_results}.

As shown in Table~\ref{tab:swe_results}, \prorlagent consistently improves performance across all model sizes. Compared with SkyRL-v0~\citep{cao2025skyrl}, the gains are particularly notable for the 8B model, where \prorlagent achieves nearly a 2$\times$ improvement on SWE-Bench Verified. These results suggest that our infrastructure provides a more effective and stable foundation for RL training on software engineering agents.

\subsection{Generality Across Agent Domains}
Beyond software engineering agents, we further demonstrate the generality of \prorlagent by conduct RL training on other domains. \\
\noindent\textbf{STEM Agent.}
We further train a STEM agent designed to solve complex question-answering tasks across science, technology, engineering, and mathematics. Its primary tool is \emph{web search}, which enables retrieval of external knowledge for open-domain reasoning. In addition, the agent is equipped with the Bash and IPython tools provided by our infrastructure, allowing it to write and execute code for numerical computation and symbolic problem solving.
For the web search backend, we use Tavily. For training data, we follow the ProRL recipe~\citep{prorl2025} and use the SCP-116K dataset~\citep{lu2025scp116khighqualityproblemsolutiondataset}.

\begin{figure*}[t]
  \centering
  \begin{subfigure}[t]{0.32\textwidth}
    \centering
    \includegraphics[width=\linewidth]{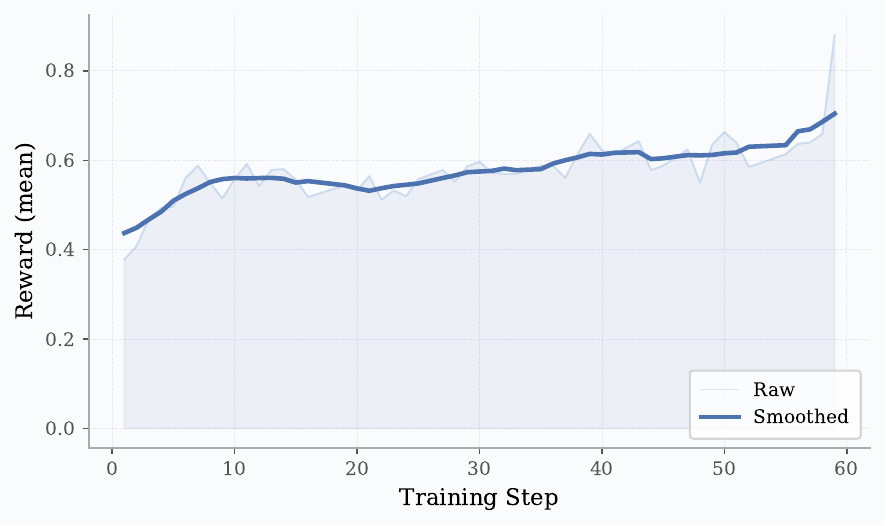}
    \caption{STEM agent.}
    \label{fig:stem_agent}
  \end{subfigure}
  \hfill
  \begin{subfigure}[t]{0.32\textwidth}
    \centering
    \includegraphics[width=\linewidth]{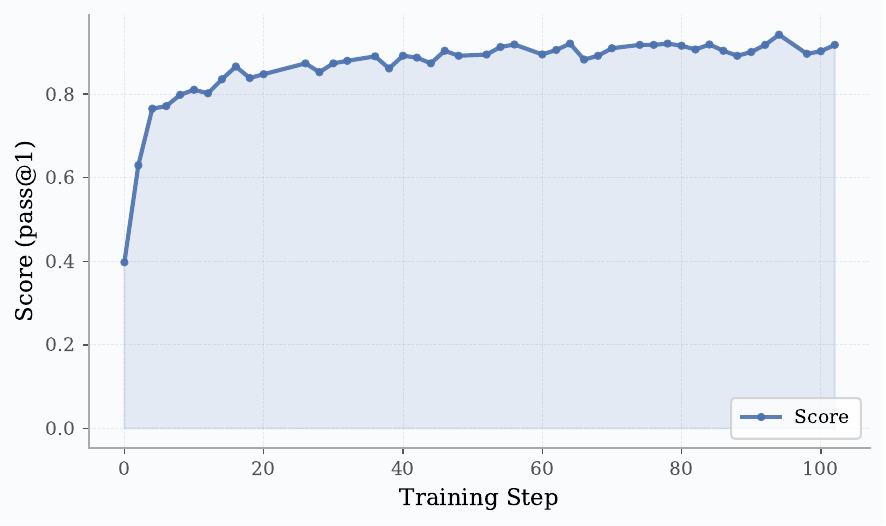}
    \caption{Math agent.}
    \label{fig:math_agent}
  \end{subfigure}
  \hfill
  \begin{subfigure}[t]{0.32\textwidth}
    \centering
    \includegraphics[width=\linewidth]{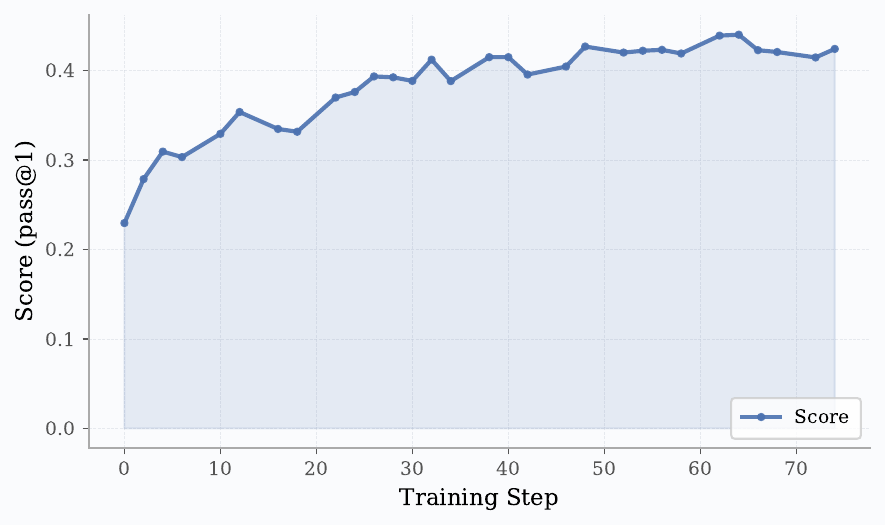}
    \caption{Code agent.}
    \label{fig:code_agent}
  \end{subfigure}
  \caption{Training curves for \prorlagent across three agent domains. From left to right: mean reward during RL training of the STEM agent, Pass@1 on AMC during RL training of the math agent, and Pass@1 on Codeforces during RL training of the code agent. All three curves show steady improvement during training, demonstrating the generality of \prorlagent beyond software engineering tasks.}
  \label{fig:other_agents}
\end{figure*}

As shown in \Cref{fig:stem_agent}, the mean reward increases steadily throughout RL training, rising from approximately 0.2 to around 0.65 after 60 training steps. The smoothed curve maintains a clear upward trend without signs of saturation, suggesting that additional training may lead to further gains. These results demonstrate that \prorlagent extends naturally beyond software engineering tasks, requiring only appropriate tool configurations and reward designs for new domains.

\noindent\textbf{Math Agent.} 
We also train a math agent to solve mathematical problems. Following ProRL~\citep{prorl2025} we use DeepScaleR~\citep{deepscaler2025} data for training and further instruct models to use tools to solve and verify its own answers. Its primary tool is \emph{IPython execution}, which provides a full computational environment with preloaded libraries such as NumPy, SciPy, and SymPy for numerical analysis and symbolic manipulation. In addition, the agent is equipped with a \texttt{think} tool for explicit planning, enabling it to decompose complex problems, devise solution strategies, and iteratively verify answers through computation. The execution backend is implemented with an IPython kernel with pre-installed scientific libraries provided by our infrastructures.

As shown in \Cref{fig:math_agent}, the Pass@1 performance on AMC improves steadily during RL training, increasing from 0.4 to approximately 0.9. The relatively low initial performance reflects the fact that the base model is not yet proficient at solving mathematical problems through simple tool use. Through RL training with \prorlagent, agent learns to effectively leverage external tools for mathematical reasoning and achieves substantial performance gains

\noindent\textbf{Code Agent.} 
We also train a code agent for program synthesis tasks. Following ProRL~\citep{prorl2025} we use Eurus-2-RL-Data~\citep{yuan2024implicitprm} as the training data and evaluate on the testing split of Codeforces. The primary tool is \emph{file editing via \texttt{str\_replace\_editor}}, which enables precise modification of source code in a dedicated \texttt{/workspace/solution.py} file. In addition, the agent is equipped with Bash execution for running test scripts and IPython tools for rapid prototyping, allowing it to iteratively develop, test, and debug solutions.
We adopt a test-driven training setup in which the agent writes verification scripts and validates outputs against expected results provided together with the problem statement. We explicitly instruct the model to verify candidate solutions with tests before submission. For reward computation, we extract the final solution from \texttt{/workspace/solution.py} and evaluate it using hidden test cases.

As shown in \Cref{fig:code_agent}, the Pass@1 performance on Codeforces improves steadily during RL training, increasing from 0.23 to approximately 0.42. Similar to the math agent, the base model initially struggles with effective use of the \texttt{str\_replace\_editor} tool and test-based verification. RL training substantially improves these capabilities, demonstrating that \prorlagent can effectively learn code generation through tool use.

\begin{table}[htb]
\centering
\small
\setlength{\tabcolsep}{5pt}
\renewcommand{\arraystretch}{1.08}
\caption{Ablation study of the proposed system components. Action Time denotes the average time required to execute shell-command actions. Each component improves rollout throughput, either by increasing GPU utilization or by reducing action execution time.}
\begin{tabular*}{\linewidth}{@{\extracolsep{\fill}}cccccc@{}}
\toprule
\textbf{Load Balancing} & \textbf{Efficient Bash}  & \textbf{Stale Job Cleanup} & \textbf{Action Time (s)} & \textbf{GPU Util (\%)} & \textbf{Throughput (instance/sec)} \\
\midrule 
\rowcolor{gray!30} \checkmark &  \checkmark   &  \checkmark  &      0.42       &          78 &    0.37        \\
     &  \checkmark   &  \checkmark  &       0.42      &   42        &       0.25     \\
          \checkmark &     &  \checkmark  &     0.78        &    68       &    0.29        \\
               \checkmark &  \checkmark   &   &        0.42     &          65 &    0.30        \\
     
   \bottomrule
\end{tabular*}
\label{tab: ablation}
\end{table}

\subsection{System Analysis}

\subsubsection{Scalability Across Compute Nodes}

\begin{figure}[t]
  \centering
  \includegraphics[width=0.55\textwidth]{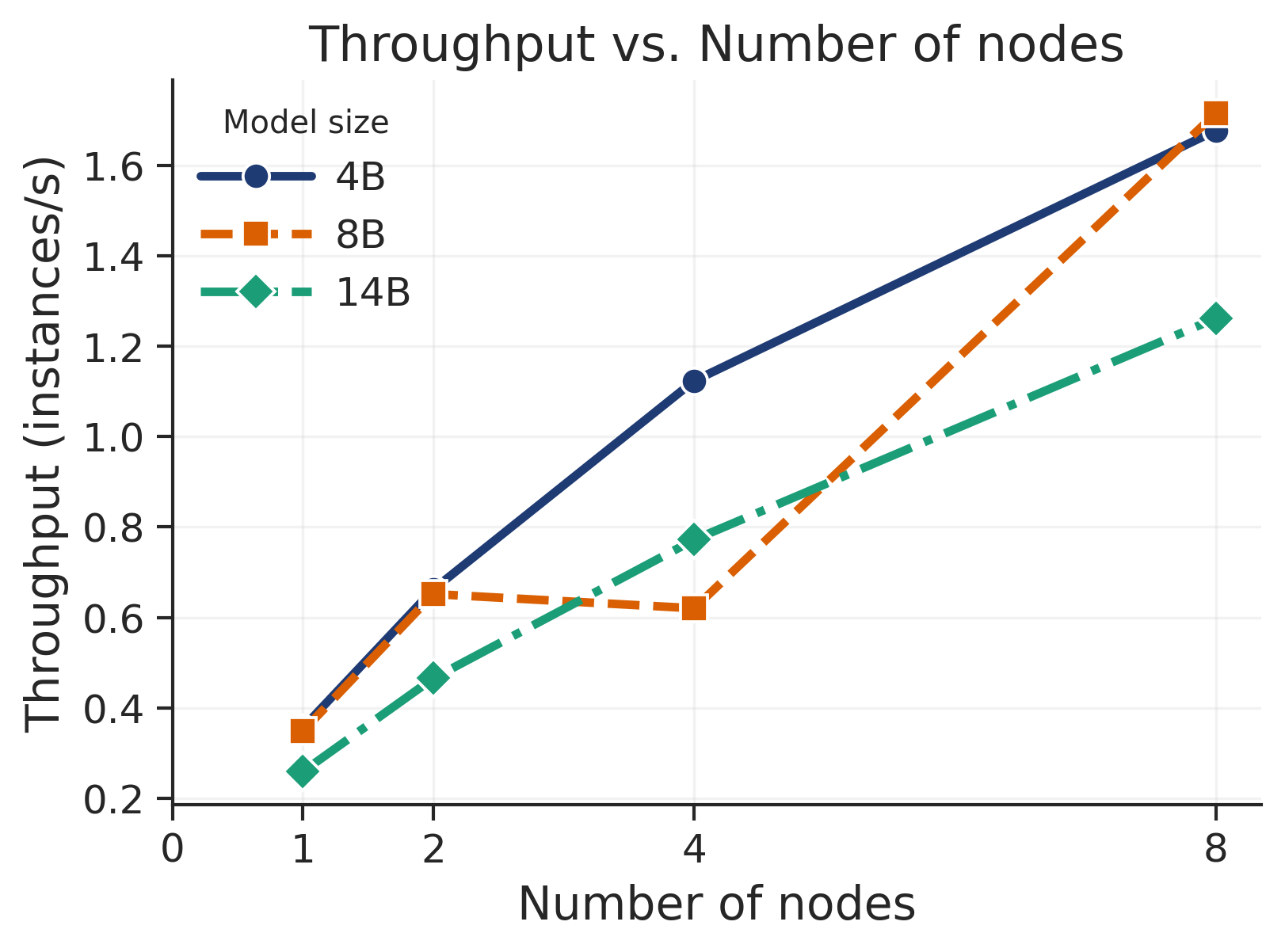}
\caption{Rollout throughput (instances/sec) on software engineering tasks versus the number of compute nodes. The near-linear increase in throughput demonstrates that \prorlagent scales efficiently with additional compute resources.}
  \label{fig:throughput}
\end{figure}

To evaluate the scalability of \prorlagent, we measure rollout throughput (instances per second) on software engineering tasks as the number of compute nodes increases. The results are shown in Fig.~\ref{fig:throughput}.

As shown in Fig.~\ref{fig:throughput}, throughput increases nearly linearly with the number of nodes, indicating that \prorlagent can effectively leverage additional compute resources with minimal scaling overhead. This scalability is particularly valuable for RL training, where efficient rollout generation is often the main system bottleneck and directly affects overall training efficiency.
\subsubsection{Component Ablations}
We then conducted ablation experiments to evaluate the effectiveness of key components of the ProRL Agent including Load Balancing (LB), Efficient Bash (EB), Stale job Cleanup (SC). 
Specifically, we measure the rollout throughput of DAPO training on Qwen3-14B-Instruct-2507 using 8 H100 GPUs, with each component removed in turn. For the variant without Load Balancing, we use a simple baseline assignment strategy that distributes an equal 
number of instances to each LLM server. For the variant without Efficient Bash, we replace our optimized implementation with the original Bash implementation from OpenHands~\citep{wang2024openhands}. For the variant without Stale Job Cleanup, we wait for all jobs to finish before proceeding.
The results in \Cref{tab: ablation} show that each proposed component contributes to higher rollout throughput during DAPO training. In particular, Load Balancing and Stale Job Cleanup improve throughput by increasing GPU utilization, while Efficient Bash improves throughput by reducing action execution time.

\section{Conclusion}

In this work, we introduce \textsc{ProRL Agent}, a open-source scalable rollout infrastructure for HPC-native multi-turn agent training. 
By separating the entire rollout lifecycle from policy training, \textsc{ProRL Agent} improves modularity, scalability, and deployability for agent RL. Experiments across software engineering, STEM, math, and code agents demonstrate effective end-to-end RL training, with strong performance gains across multiple model scales. We release \textsc{ProRL Agent} as open source and as part of NVIDIA NeMo Gym, and leave richer environments and improved cluster-scale robustness to future work.

\clearpage
\newpage
\bibliographystyle{plainnat}
\bibliography{reference}

%%%%%%%%%%%%%%%%%%%%%%%%%%%%%%%%%%%%%%%%%%%%%%%%%%%%%%%%%%%%%%%%%%%%%%%%%%%%%%%
%%%%%%%%%%%%%%%%%%%%%%%%%%%%%%%%%%%%%%%%%%%%%%%%%%%%%%%%%%%%%%%%%%%%%%%%%%%%%%%
% APPENDIX
%%%%%%%%%%%%%%%%%%%%%%%%%%%%%%%%%%%%%%%%%%%%%%%%%%%%%%%%%%%%%%%%%%%%%%%%%%%%%%%
%%%%%%%%%%%%%%%%%%%%%%%%%%%%%%%%%%%%%%%%%%%%%%%%%%%%%%%%%%%%%%%%%%%%%%%%%%%%%%%
\newpage
\appendix
\onecolumn
\section{Appendix}

Here, we provide a detailed architectural analysis of existing agent RL infrastructures, accompanied by illustrative diagrams.

\begin{figure}[h]
  \centering
  \includegraphics[width=0.9\linewidth]{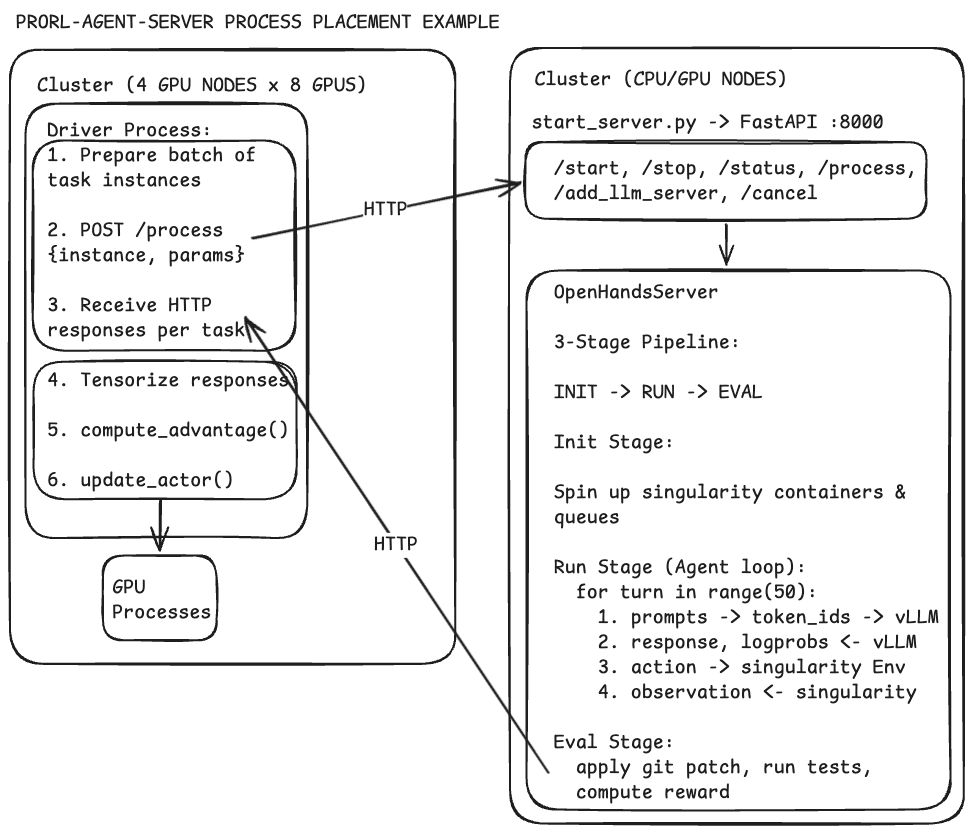}
  \caption{\prorlagent separates the full agentic rollout lifecycle, spanning environment management to reward computation, from GPU-intensive training, thereby decoupling I/O-intensive rollout from training.}
  \label{fig:prorl_agent}
\end{figure}

\begin{figure}[h]
  \centering
  \includegraphics[width=1\linewidth]{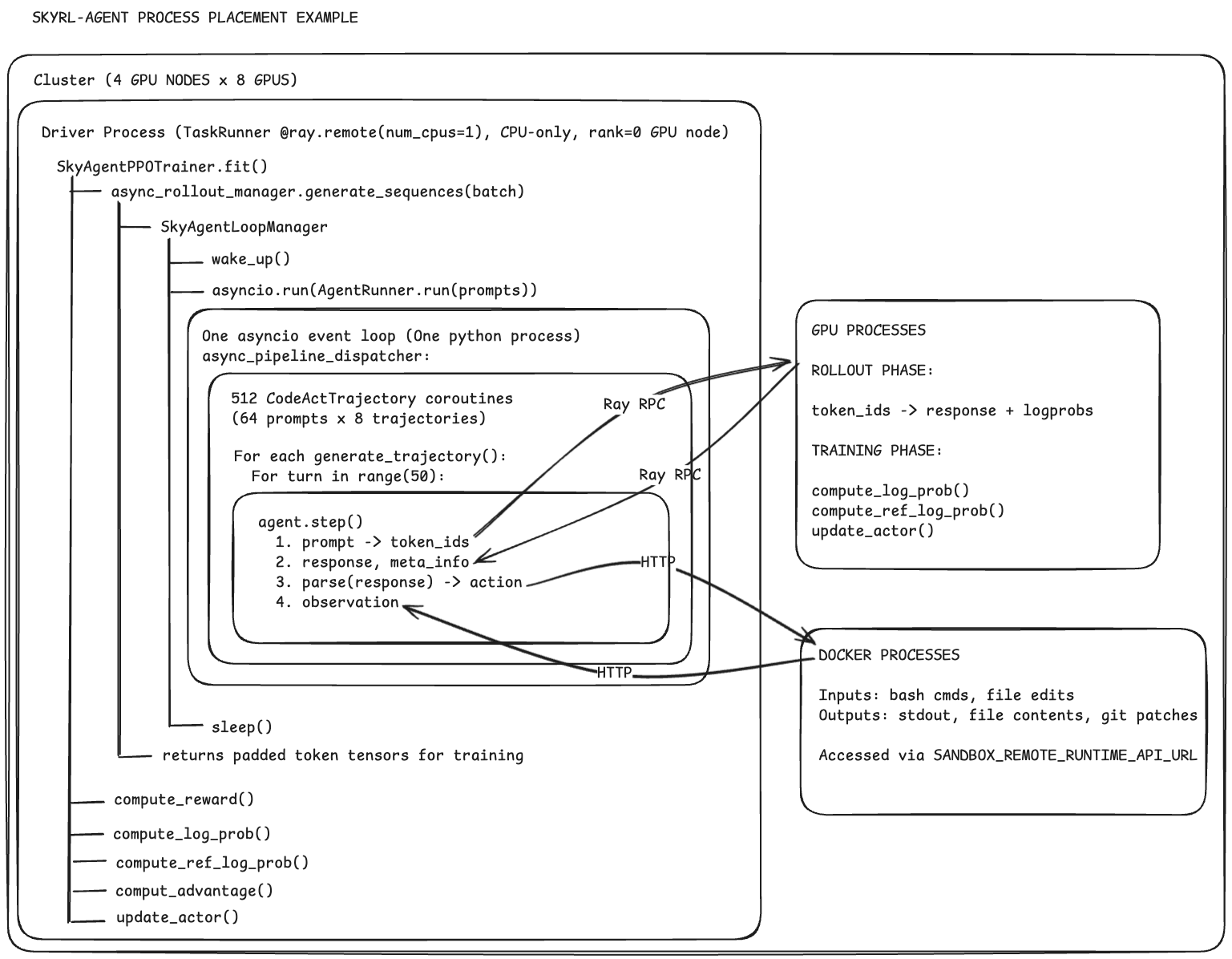}
\caption{
\textbf{SkyRL-Agent.} The training driver runs concurrent trajectory-generation coroutines on a single CPU process. It controls the multi-turn agent loop, queries a remote vLLM server for inference, and interacts with remote environment containers for execution. Although inference and environment execution are offloaded, rollout control remains inside the training driver.}
  \label{fig:skyrl_agent}
\end{figure}

\begin{figure}[h]
  \centering
  \includegraphics[width=0.55\textwidth]{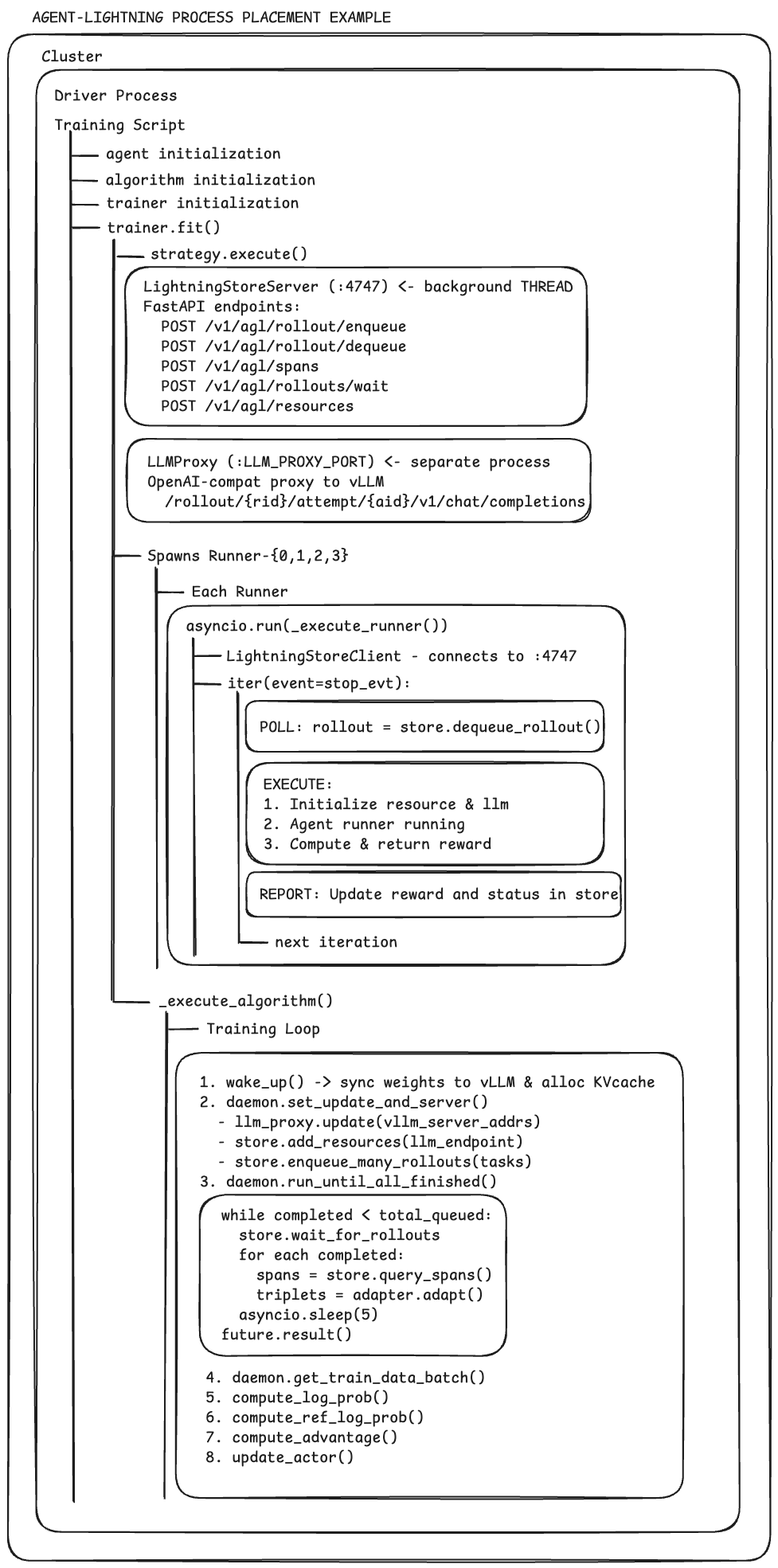}
\caption{\textbf{Agent Lightning.} Agent Lightning places the training loop, the \texttt{LightningStoreServer}, and all rollout workers within a single process tree. The store runs as a background thread, while rollout workers are spawned as child processes from the trainer. As a result, rollout does not have an independent service lifecycle: if the training process terminates, the store also stops and the rollout workers are disrupted. Thus, rollout remains managed within the training stack rather than being cleanly decoupled.}
  \label{fig:agent_lightning}
\end{figure}

\begin{figure}[h]
  \centering
  \includegraphics[width=1\textwidth]{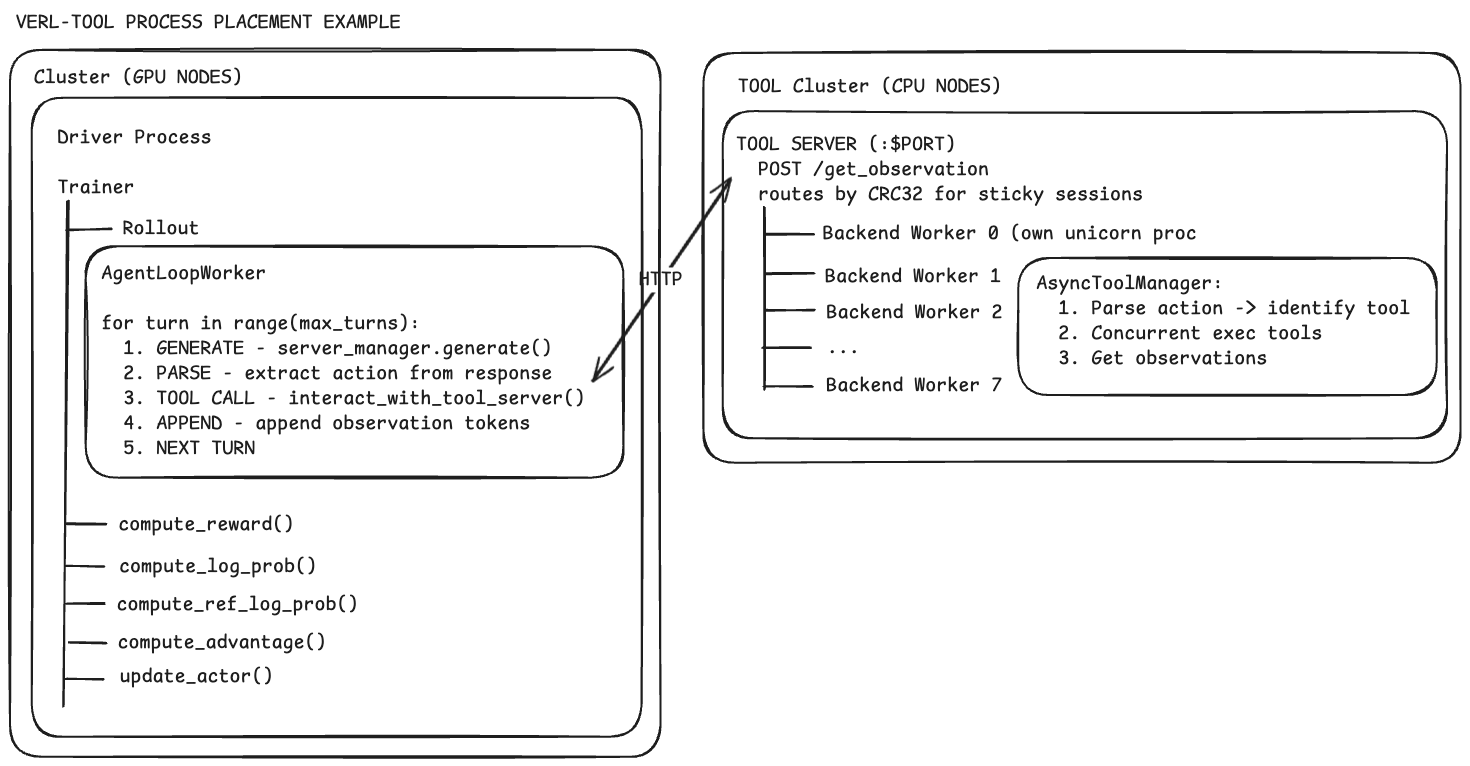}
\caption{\textbf{VeRL-Tool.} VeRL-Tool extends the standard veRL trainer to support multi-turn agent rollouts. The training system manages the agent loop and trajectory collection, while tool execution is offloaded to a separate CPU-based environment service. In this design, rollout control remains inside the trainer.}
  \label{fig:verl_tool}
\end{figure}

\begin{figure}[h]
  \centering
  \includegraphics[width=0.7\textwidth]{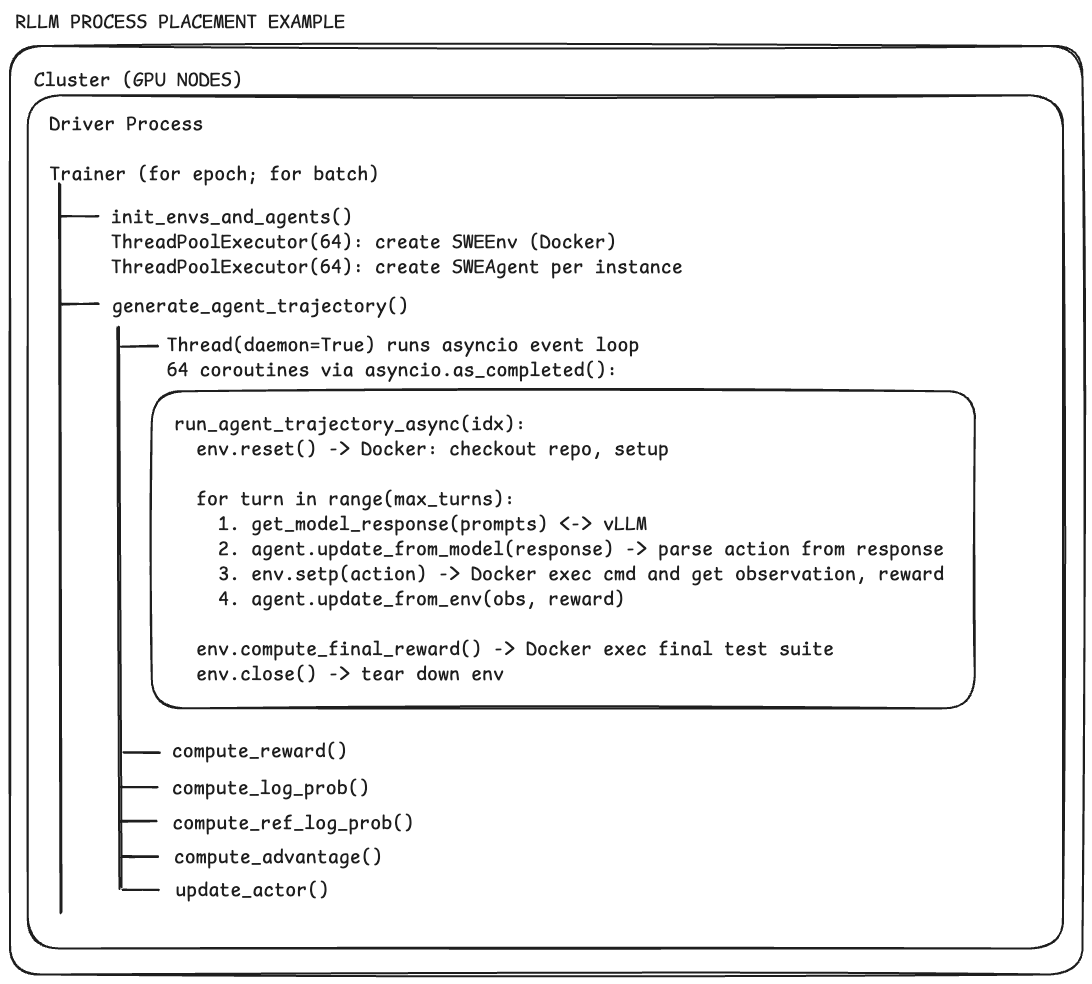}
\caption{\textbf{rLLM: rollout embedded in a monolithic training driver.} rLLM is built on a heavily modified fork of veRL. The agent loop, environment management, and trajectory orchestration all reside within a single driver process. There is no independent rollout service, no persistent trajectory buffer, and no possibility of the rollout surviving independently of the training driver. The full rollout lifecycle remains tightly coupled with the training stack.}
  \label{fig:rllm}
\end{figure}

\begin{figure}[h]
  \centering
  \includegraphics[width=0.7\textwidth]{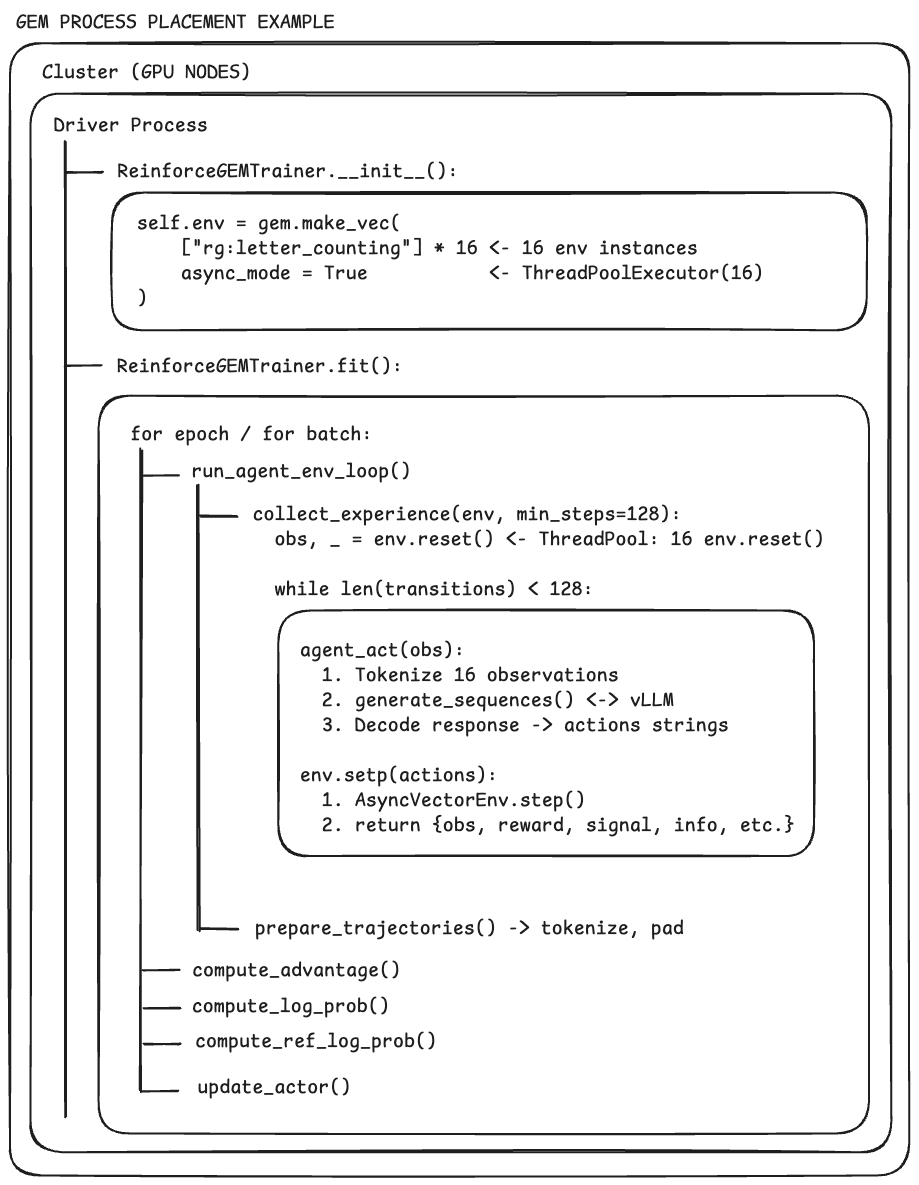}
\caption{\textbf{GEM.} GEM keeps environment execution inside the training process. Environments are instantiated as in-memory Python objects, and environment stepping is performed through direct \texttt{env.step()} calls, with parallelism provided only by \texttt{ThreadPoolExecutor} threads in the same address space. A single driver process orchestrates both rollout and training, while GPU workers are accessed remotely via Ray RPC. As a result, the environment and rollout lifecycle remain fully embedded in the training stack.}
  \label{fig:gem}
\end{figure}
%%%%%%%%%%%%%%%%%%%%%%%%%%%%%%%%%%%%%%%%%%%%%%%%%%%%%%%%%%%%%%%%%%%%%%%%%%%%%%%
%%%%%%%%%%%%%%%%%%%%%%%%%%%%%%%%%%%%%%%%%%%%%%%%%%%%%%%%%%%%%%%%%%%%%%%%%%%%%%%

\end{document}